\documentclass[runningheads]{llncs}
\usepackage[T1]{fontenc}
\usepackage{amsmath}
\usepackage{graphicx}
\usepackage{hyperref}
\usepackage{tikz}
\usepackage{pgfplots}
\pgfplotsset{compat=1.18}
\usepgfplotslibrary{groupplots}
\usetikzlibrary{shapes.geometric,calc}
\usepackage{bm}
\usepackage{float}
\usepackage{longtable}
\usepackage[numbers,square]{natbib}
\usepackage{booktabs}
\usepackage{gensymb}
\usepackage{caption}
\usepackage{subcaption}
\usepackage{threeparttable}
\usepackage{array}
\begin{document}
\title{Exploring \texorpdfstring{\textit{napping}}{napping} paradigm for \texorpdfstring{\\}{} Recurrent Spiking Neural Networks}
\titlerunning{Exploring \textit{napping} paradigm for Recurrent Spiking Neural Networks}
%
\author{Andreas Massey\inst{1}\orcidID{0009-0002-5300-9642} \and
Stefano Nichele\inst{2}\orcidID{0000-0003-4696-9872} \and
Aliaksandr Hubin\inst{1}\orcidID{0000-0002-3244-6571}
}
\authorrunning{A. Massey et al.}
%
\institute{The Norwegian University of Life Sciences, \\Elizabeth Stephansens v. 15, 1433 Ås, Norway \and
Østfold University of Applied Sciences, \\B R A Veien 4, 1757 Halden, Norway}


\maketitle              
\begin{abstract}
Biological organisms minimize free energy by balancing two competing demands on their internal world model: it must be accurate enough to predict sensory input, yet simple enough to generalize beyond it. Two mechanisms regulate this balance offline: sleep reduces complexity through gradual synaptic downscaling, while stochastic noise attenuates precision, relaxing the constraint sensory input imposes on synaptic reorganization. Engineered Spiking Neural Networks (SNNs) leave this balance unaddressed, favoring instantaneous, noiseless weight normalization instead. This paper investigates the hypothesis that a biologically inspired micro-sleep paradigm \textit{napping}---combining proportional weight scaling with continuous stochastic membrane activity---can replicate the stability of normalization while shedding model complexity. We evaluate this in an unsupervised recurrent SNN trained via trace-based spike-timing-dependent plasticity (STDP) on Gabor-preprocessed MNIST. We tune napping across three regularization regimes by sweeping its duration and membrane noise level, then compare the best configuration against weight normalization. Across all three regimes, well-tuned napping matches the accuracy of normalization: accuracy peaks at brief durations and low noise, then declines monotonically as either grows. Clustering diverges, with the strongest geometric separation arising at longer durations and higher noise---the two terms of free energy pulling apart, accuracy rewarding data fit and structure rewarding the simpler representation that gradual, noisy downscaling induces. This gain carries a simulation cost normalization avoids, so napping is most compelling where representational structure, rather than raw classification efficiency, is the priority.

\end{abstract}

\section{Introduction}
Biological neural circuits sustain accurate models of a noisy, ever-changing world without letting those models grow too complex to generalize---a balance engineered spiking neural networks (SNNs) have yet to strike. Two mechanisms regulate this balance offline. First, \textit{sleep}: under the synaptic homeostasis hypothesis (SHy), gradual, activity-weighted synaptic downscaling bounds weight growth and consolidates learned structure, reducing model complexity \cite{tononi_sleep_2014}. Second, \textit{noise}: stochastic fluctuations in membrane potentials, synaptic release, and background activity attenuate the precision of sensory drive, preventing rigid weight configurations and sustaining representational diversity. The most efficient SNNs employ event-encoded information enabling high performance on neuromorphic hardware \cite{su_deep_2023}, with orders-of-magnitude reductions in power consumption over dense architectures \cite{isik_accelerating_2024}, yet their dominant learning mechanisms leverage neither principle. Surrogate gradient methods yield strong performance \cite{massey_sleep-based_2026} but are incompatible with neuromorphic hardware, undermining the efficiency advantage SNNs promise. The canonical local alternative, STDP, is prone to homeostatic instability, particularly in recurrent architectures where uncontrolled weight growth produces pathological attractor states that fail to generalize.   

\subsection{The Problem with Learning}
STDP is the canonical local learning rule for SNNs \cite{song_competitive_2000}. Weight modifications depend on millisecond-scale timing differences between pre- and postsynaptic spikes: preceding presynaptic activity strengthens weights; reversed order weakens them. This causality-respecting rule implements credit assignment through temporal adjacency and spike coincidence detection \cite{jayabal_experience_2024,chauhan_emergence_2018}.

Biological STDP operates in an inherently noisy environment; stochastic variability in spike timing regularizes weight updates and partially offsets STDP's intrinsic tendency toward runaway potentiation \cite{eppler_representational_2026}. Engineered SNNs suppress this variability in pursuit of deterministic reproducibility, inadvertently removing a natural check on model complexity.

In recurrent SNN architectures---where lateral connectivity is essential for sustaining internal state---STDP introduces severe instability: positive feedback amplifying weight perturbations, driving three intertwined failure modes: (i) \textbf{weight saturation}, where unrestricted Hebbian potentiation drives weights toward unbounded values or clipping limits \cite{gilson_stdp_2010}; (ii) \textbf{representational collapse}, where weight coupling induces spurious correlations that erode discriminative structure \cite{akil_synaptic_2020}; and (iii) \textbf{catastrophic forgetting}, where sequential learning overwrites previously consolidated representations \cite{tadros_sleep-like_2022}. In free-energy terms these are failures to constrain model complexity.

\subsection{Napping as Free-Energy Solution}
\label{sec::intro:homeostasis}
The most common computational---but not biologically inspired---remedy is \textit{weight normalization}: synaptic weights rescaled at regular intervals by a factor that drives the weight distribution toward a regime norm---either the weights at initialization (hereafter the \textit{static} regime), a per-layer mean, or a per-post-neuron mean---to prevent runaway growth \cite{carlson_biologically_2013}. While effective, normalization is applied instantaneously, operates externally to the learning process, and is entirely noiseless \cite{ioffe_batch_2015}.
The synaptic homeostasis hypothesis (SHy) posits a fundamentally different mechanism. During sleep, external inputs are suppressed, spontaneous activity drives continued STDP, and weights decay toward a target gradually over the sleep phase \cite{tononi_sleep_2003}. This process is inherently noisy since spontaneous firing is stochastic and the gradual downscaling allows continued reorganization of weight structure. Sleep is therefore not merely a homeostatic shrinkage of weights but an active consolidation phase. Under a free-energy view, this is the complexity term in action: sleep reduces model complexity by downscaling synapses. The stochasticity of spontaneous firing plays the role of noise, lowering precision so the model can reorganize \citep{hobson_waking_2012}.

Building on a prior SNN sleep model \cite{massey_sleep-based_2026}, we identify a limitation specific to weight decay toward a fixed absolute target: it erodes the \textit{relative} structure of the weight distribution. As weights converge toward the same value, learned differences between strong and weak synapses are eroded regardless of their functional importance — a pathology absent from normalization, which rescales by a common ratio and thus preserves relative magnitudes.

Napping---brief proportional weight scaling embedded within sleep dynamics---represents a candidate resolution to this pathology. By scaling weights by a common ratio rather than decaying toward a fixed target, relative synaptic magnitudes are preserved; by applying this scaling gradually and under stochastic activity, the consolidation properties of sleep are nominally retained. This paper tests whether napping can in practice reconcile the stability of normalization with the noisy consolidation of biological sleep. 

\section{Methods}
We train an unsupervised three-layer recurrent SNN---a Poisson-encoded input layer, a recurrent excitatory layer, and a lateral inhibitory population \cite{zenke_diverse_2015}---using exclusively local learning rules. MNIST images \cite{yann_lecun_mnist_2010} are preprocessed through Gabor filters at four orientations, approximating V1 orientation-selective receptive fields \cite{daugman_uncertainty_1985}. Recurrent excitation is retained to sustain the attractor dynamics necessary for stable class representations; the objective is to regularize it, not eliminate it\footnote{Thiele et al. \cite{thiele_wake-sleep_2017} successfully leveraged anti-STDP rules to \textit{unlearn} strong attractors, however, this is not the target hypothesis in the current work.}.

We sweep napping duration and membrane noise level over five seeds across three regularization regimes defined by the magnitude each rescales toward: initialization values (\textit{static}), the per-layer mean (\textit{layer}), and per post-synapse mean (\textit{neuron}). This addresses two questions: whether gradual, noisy downscaling outperforms instantaneous rescaling, and whether dynamic (layer, neuron) regimes preserve weight diversity better than the fixed static regime. 

Finally, we compare the optimal setup for napping against normalization across the regularization regimes and independent seeds.  

All code for the study is publicly available in \texttt{experiments/noise\_article} in 
 \cite{no-author_spikingneuralnetwork_nodate}.

\subsection{Neuron Model}
Each neuron is modeled using leaky integrate-and-fire (LIF) dynamics, where the membrane potential decays toward rest between spikes and rises with weighted spike input from connected neurons. To prevent runaway activity, we implement an adaptive spike threshold that rises with each emitted spike and decays exponentially back to baseline. During napping, Gaussian noise replaces structured input to drive stochastic spiking in the absence of external stimulation. The formal mathematical framework and parameter definitions are provided in \cite{massey_sleep-based_2026} and Appendices~\ref{sec::app:hyperparameters} and~\ref{sec::app_neuron}.

\subsection{Network Architecture}
\label{sec::met:network}
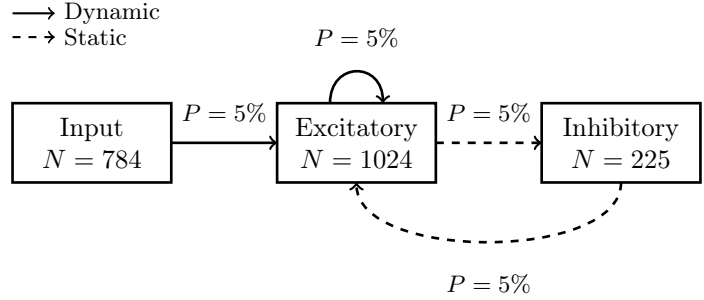
\begin{figure}[H]
\centering
\begin{tikzpicture}[scale=0.7, line width=1.0pt, font=\normalsize]

\draw (0, 0) rectangle (3, 1.5) node[midway, align=center] {Input\\ $N = 784$};

\draw (5, 0) rectangle (8, 1.5) node[midway, align=center] {Excitatory\\ $N=1024$};

\draw (10, 0) rectangle (13, 1.5) node[midway, align=center] {Inhibitory\\ $N = 225$};

\draw[->, line width=1.0pt] (3, 0.75) -- (5, 0.75) node[midway, above, yshift=0.15cm] {\small $P = 5\%$};
\draw[->, line width=1.0pt, dashed] (8, 0.75) -- (10, 0.75) node[midway, above, yshift=0.15cm] {\small $P = 5\%$};

\draw[->, line width=1.0pt] (6, 1.5) .. controls (6, 2.3) and (7, 2.3) .. (7, 1.5) node[midway, above, yshift=0.2cm] {\small $P=5\%$};

\draw[->, line width=1.0pt, dashed] (11.5, 0) .. controls (11.5, -1.5) and (6.5, -1.5) .. (6.5, 0) node[midway, below, yshift=-0.3cm] {\small $P=5\%$};

\begin{scope}[font=\footnotesize]
\draw[->, line width=0.8pt] (0, 3.2) -- (0.8, 3.2) 
    node[right] {Dynamic};
\draw[->, line width=0.8pt, dashed] (0, 2.75) -- (0.8, 2.75) 
    node[right] {Static};
\end{scope}
\end{tikzpicture}
\caption{Network architecture: three-layer SNN with feedforward input ($P=5\%$), recurrent excitation ($P=5\%$), and lateral inhibition ($P=5\%$). The solid lines denote weights that change over time, and dashed lines represent static weights.}
\label{fig:architecture}
\end{figure}

Following \cite{zenke_diverse_2015,diehl_unsupervised_2015}, the network consists of three 
layers: a Poisson spike-encoded input layer ($N_{\text{in}}=784$), a 
recurrent excitatory layer ($N_{\text{exc}}=1024$), and a lateral 
inhibitory layer ($N_{\text{inh}}=225$). Across all layers, 
connectivity is random and sparse ($P=5\%$ for all pathways); see Figure~\ref{fig:architecture}.

\subsection{Learning}
Following \cite{diehl_unsupervised_2015}, we use trace-based STDP, in
which presynaptic activity accumulates into a leaky continuous trace,
enabling periodic weight updates based on recent spike
timing. Weights are updated by
\begin{equation}
\label{eq:STDP-trace}
    \Delta w = \eta\!\left(x_{pre}-x_{tar}^l\right)(w_{max}-w)^\mu,
\end{equation}
where $\eta$ is the learning rate, $x_{tar}^l$ is the layer-mean
presynaptic trace, $w_{max}$ is the maximum weight, and $\mu$ tapers
update magnitude as weights approach $w_{max}$. Neurons exceeding
$x_{tar}^l$ are potentiated; those below are depressed, introducing
local competition within each layer. Hyperparameters are listed in
Appendix~\ref{sec::app:hyperparameters}.

\subsection{Regularization}
We compare two weight regularization strategies: standard normalization 
and the napping paradigm. Both share a common scaling 
structure but differ in how and when the regime is applied.

\subsubsection{Regularization Modes}

\paragraph{Normalization}
Weight normalization is a common strategy for maintaining synaptic 
stability in SNNs~\cite{diehl_unsupervised_2015}. Following each 
training epoch, weights are rescaled instantaneously with a scaling parameter $\rho$:
\begin{equation}
    w(t)^{\text{new}} = w(t)\rho(t).
\end{equation}

\paragraph{Napping}
Napping extends normalization by distributing the weight update across 
$d$ timesteps while simultaneously injecting noisy membrane currents and enabling continued STDP 
learning. At each napping timestep, weights are scaled by a fixed 
fraction of the total rescaling:
\begin{equation}
    w(t)^{\text{new}} = w(t)\rho^{\lambda}(t), \qquad \lambda = \frac{1}{d}.
\end{equation}
After $d$ timesteps, the cumulative effect is identical to 
instantaneous normalization, $w(t) \cdot \rho(t)^{d \cdot \frac{1}{d}} = 
w(t)\rho(t)$, but the gradual application allows the network to consolidate 
activity-driven weight structure through continued STDP and noisy 
membrane dynamics. Napping is triggered in the same regular intervals as normalization during training. Hyperparameters are given in 
Appendix~\ref{sec::app:hyperparameters}.

\subsubsection{Regularization Regimes}
Both methods are parameterized by $\rho(t)$, for which we 
define three modes: static, layer, and 
neuron. The latter two are dynamic, adapting $\rho(t)$ to 
current network activity at time $t$. In all cases, $w_{ij}(0)$\footnote{$w_{ij}(t)\geq 0 $ for all $
i$,$j$, and $t$, by design since these weights are dynamic and excitatory, see Section~\ref{sec::met:network}.} denotes the 
weight from neuron $j$ to neuron $i$ at initialization.

\paragraph{Static}

The static regime applies a fixed ratio $\rho(t)=1$ per layer, driving weights back toward their initialization values; since initial weights are drawn per-layer, this implicitly defines a layer-specific absolute target. As discussed in Section~\ref{sec::intro:homeostasis}, this collapses weight variance and is therefore included as a baseline reference.

\paragraph{Layer}
The layer-wise regime computes $\rho(t)$ as the ratio of initial to 
current total weight magnitude within layer $l$, preserving relative 
inter-layer structure:
\begin{equation}
    \rho(t) = \frac{\sum_{i \in l}\sum_{j} w_{ij}(0)}
                {\sum_{i \in l}\sum_{j} w_{ij}(t)}.
\end{equation}

\paragraph{Neuron}
The neuron-wise regime computes $\rho(t)$ per post-synaptic neuron $i$ 
using only its local incoming weights, preserving neuron-level relative 
weight structure while allowing heterogeneous scaling across the 
network:
\begin{equation}
    \rho(t) = \frac{\sum_{j} w_{ij}(0)}
                {\sum_{j} w_{ij}(t)}.
\end{equation}

\begin{figure}
    \centering
    \includegraphics[width=1.0\linewidth]{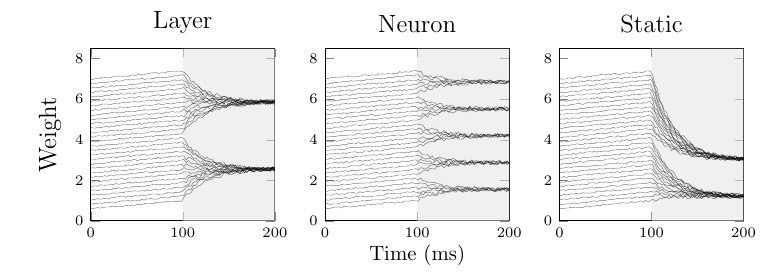}
    \caption{Effect of the three regularization regimes on synaptic weights.
        Unshaded is wake phase and shaded is napping session. \textbf{Layer}: each population converges to its layer mean.
        \textbf{Neuron}: incoming weights of each post-synaptic neuron
        converge to that neuron's mean. \textbf{Static}: both populations decay
        toward fixed initialization-derived targets.}
    \label{fig::reg_modes}
\end{figure}

\subsection{Data}
We trained the model on the MNIST dataset, which consists of $60\,000$ training images and $10\,000$ test images of handwritten digits (0--9), each of size $28\times28$ pixels, with one \textit{run} corresponding to a single epoch over the full training set. Prior to spike encoding, each image is filtered with four Gabor kernels oriented at $0\degree$, $90\degree$, $45\degree$, and $135\degree$, approximating the orientation-selective receptive fields of V1 simple cells~\cite{daugman_uncertainty_1985}; the 2D Gabor is defined further in Appendix~\ref{sec::app_data}. Each composite input value is then treated as the firing probability of a Poisson neuron, producing a binary spike train over $T=350$ ms with a maximum spiking rate of $90$ Hz, which yields a spike matrix of shape $T\times N_{\text{in}}$ per sample.

\subsection{Evaluation}
\label{sec::eval}
The model was evaluated with two distinct measurements: \textit{accuracy} and \textit{clustering}. Both methods leverage mean spike rates per sample per neuron, which is subsequently standardized to Z-scores. We then project this data from \textit{N}-dimensional space into a PCA subspace retaining up to 15 components\footnote{A conservative upper bound chosen to ensure the decomposition remains well-defined across all experimental configurations, rather than a tuned hyperparameter.}.   

\subsubsection{Accuracy}
To compute the accuracy, we train a $L_1$-regularized multinomial logistic regression (MLR) on the PCA components. The trained classifiers can then predict classes on the unseen test data processed by the network. 

\subsubsection{Clustering}
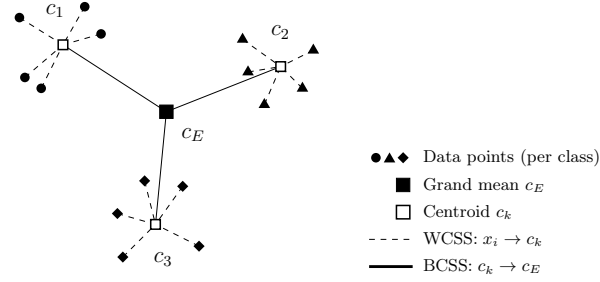
\begin{figure}[h]
\centering
\scalebox{0.72}{%
\begin{tikzpicture}

\tikzset{
  c1pt/.style={circle, fill=black, minimum size=5pt, inner sep=0pt},
  c2pt/.style={regular polygon, regular polygon sides=3,
               fill=black, minimum size=7pt, inner sep=0pt},
  c3pt/.style={diamond, fill=black, minimum size=6pt, inner sep=0pt},
  centmark/.style={rectangle, fill=white, draw=black,
                   line width=0.8pt, minimum size=5pt, inner sep=0pt},
}

\coordinate (c1) at (1.5, 4.2);
\coordinate (c2) at (5.5, 3.8);
\coordinate (c3) at (3.2, 0.9);
\coordinate (gm) at (3.4, 2.97);

\foreach \x/\y in {0.7/4.7, 1.1/3.4, 2.2/4.4, 1.9/4.9, 0.8/3.6}
  { \draw[dashed, thin] (\x,\y) -- (c1); }
\foreach \x/\y in {4.8/4.3, 5.2/3.1, 6.1/4.1, 5.9/3.4, 4.9/3.7}
  { \draw[dashed, thin] (\x,\y) -- (c2); }
\foreach \x/\y in {2.6/0.3, 3.0/1.7, 4.0/0.5, 3.7/1.6, 2.5/1.1}
  { \draw[dashed, thin] (\x,\y) -- (c3); }

\draw[line width=0.5pt] (c1) -- (gm);
\draw[line width=0.5pt] (c2) -- (gm);
\draw[line width=0.5pt] (c3) -- (gm);

\foreach \x/\y in {0.7/4.7, 1.1/3.4, 2.2/4.4, 1.9/4.9, 0.8/3.6}
  { \node[c1pt] at (\x,\y) {}; }
\foreach \x/\y in {4.8/4.3, 5.2/3.1, 6.1/4.1, 5.9/3.4, 4.9/3.7}
  { \node[c2pt] at (\x,\y) {}; }
\foreach \x/\y in {2.6/0.3, 3.0/1.7, 4.0/0.5, 3.7/1.6, 2.5/1.1}
  { \node[c3pt] at (\x,\y) {}; }

\node[centmark] at (c1) {};
\node[font=\large, xshift=-4pt, yshift=18pt] at (c1) {$c_1$};
\node[centmark] at (c2) {};
\node[font=\large, xshift=0pt, yshift=18pt] at (c2) {$c_2$};
\node[centmark] at (c3) {};
\node[font=\large, xshift=4pt, yshift=-18pt] at (c3) {$c_3$};

\node[rectangle, fill=black, draw=black,
      minimum size=7pt, inner sep=0pt] at (gm) {};
\node[font=\large, xshift=14pt, yshift=-12pt] at (gm) {$c_E$};


\coordinate (LA) at (7.25, 2.125);   
\node[c1pt] at (LA) {};
\coordinate (LB) at (7.5, 2.1);
\node[c2pt] at (LB) {};
\coordinate (LC) at (7.75, 2.125);
\node[c3pt] at (LC) {};
\node[font=\footnotesize, anchor=west] at (8, 2.125)
  {Data points (per class)};

\coordinate (LD) at (7.75, 1.125);   
\node[centmark, minimum size=7pt] at (LD) {};
\node[font=\footnotesize, anchor=west] at (8.0, 1.125) {Centroid $c_k$};

\coordinate (LE) at (7.75, 1.625);   
\node[rectangle, fill=black, draw=black,
      minimum size=7pt, inner sep=0pt] at (7.75, 1.625) {};
\node[font=\footnotesize, anchor=west] at (8.0, 1.625) {Grand mean $c_E$};

\coordinate (LF) at (7.125, 0.625);
\draw[dashed, thin] (LF) -- ($(LF)+(0.8,0)$);
\node[font=\footnotesize, anchor=west] at (8.0, 0.625)
  {WCSS:\;$x_i \to c_k$};

\coordinate (LG) at (7.125, 0.125);
\draw[line width=1.5pt] (LG) -- ($(LG)+(0.8,0)$);
\node[font=\footnotesize, anchor=west] at (8.0, 0.125)
  {BCSS:\;$c_k \to c_E$};

\end{tikzpicture}
}
\caption{Illustration of the Calinski--Harabasz (CH) index. Dashed lines
indicate within-cluster distances (WCSS) from each point $x_i$ to its class
centroid $c_k$; solid lines indicate between-cluster distances (BCSS) from
each centroid to the grand mean $c_E$.}
\label{fig:ch_index}
\end{figure}

For clustering, we use the clustering method Calinski-Harabasz (CH) index \cite{tadeusz_calinski_dendrite_1974} on the PCA-transformed subspace
\begin{equation}
    \text{CH} = \frac{\frac{\text{BCSS}}{k-1}}{\frac{\text{WCSS}}{n-k}},
\end{equation}
where BCSS is the between-clusters sums-of-squares, WCSS is the within-cluster sums-of-squares, $n$ is the number of samples and $k$ is the number of centroids. We illustrate the clustering index in Figure~\ref{fig:ch_index}.


\subsection{Experimental Design}
\label{sec::met_design_ablation}
Our analysis proceeds in two phases. In the first, we \textit{tune} napping by independently varying its two distinguishing properties---scaling duration and stochastic membrane activity---across all three regularization regimes, isolating their functional contribution. In the second, we \textit{compare} the resulting napping behavior directly against weight normalization. Each phase is statistically analyzed with its own generalized linear mixed-effects model (GLMM), pre-specified so that every fixed effect encodes a distinct a priori hypothesis; since no post-hoc pairwise contrasts are performed, $p$-values are reported unadjusted.

\paragraph{Phase 1: Sweep.}
We sweep napping duration and noise level across regularization regimes and five repetitions per setting with independent seeds:
\begin{equation}
    \text{Napping duration (7)} \times \text{Noise level (6)}
    \times \text{Reg. regime (3)} \times \text{Seed (5)}.
\end{equation}
We model each outcome metric with a GLMM: accuracy, bounded in $(0,1)$, with a Beta family and logit link; clustering score, strictly positive and right-skewed, with a Gamma family and log link. Napping duration and noise level are log-transformed to account for their non-linear parameter ranges and enter as continuous predictors:
\begin{align}
y^{\text{acc}}_{i} &\sim \mathrm{Beta}\!\left(\mu^{\text{acc}}_{i},\phi\right), \quad
y^{\text{clust}}_{i} \sim \mathrm{Gamma}\!\left(\mu^{\text{clust}}_{i},\phi\right), \\
g\!\left(\mu^{m}_{i}\right)
&= \beta_0
+ \beta_1 s_i
+ \beta_2 n_i
+ \beta_3 I^{\text{layer}}_i
+ \beta_4 I^{\text{neuron}}_i
+ \beta_5\, s_i n_i \nonumber \\
&\quad
+ \beta_6\, s_i I^{\text{layer}}_i
+ \beta_7\, s_i I^{\text{neuron}}_i
+ \beta_8\, n_i I^{\text{layer}}_i
+ \beta_9\, n_i I^{\text{neuron}}_i \nonumber \\
&\quad
+ \beta_{10}\, s_i n_i I^{\text{layer}}_i
+ \beta_{11}\, s_i n_i I^{\text{neuron}}_i
+ b_{\text{seed}(i)},
\end{align}
where $g(\cdot)$ is the logit link for accuracy and the log link for clustering; 
$s_i = \log(\text{duration}_i)$ and $n_i = \log(\text{noise}_i)$; 
$I^{\text{layer}}_i$ and 
$I^{\text{neuron}}_i$ are indicators for the regularization regimes (with ``static'' as the reference level); 
$b_{\text{seed}(i)} \sim \mathcal{N}(0,\sigma^2)$ is a per-seed random intercept; 
and $\phi$ are dispersion parameters.

\paragraph{Phase 2: Comparison.}
To estimate the effect of napping relative to normalization, we fit a second GLMM for comparing accuracy and clustering with respect to regularization types (optimal setting of napping\footnote{Accuracy-optimal: duration $= 1$, noise $\in \{5, 10\}$ across regimes. Clustering-optimal: duration $\in \{1, 100, 300\}$, noise $\in \{1, 2.5, 100\}$. Static excluded, see the Appendix Table~\ref{tab::app:glmm:p2:optima}.}, with regime (layer, neuron, or static) and a per-seed random intercept:
\begin{align}
y^{\text{acc}}_{i} &\sim \mathrm{Beta}\!\left(\mu^{\text{acc}}_{i}, \phi\right), \quad
    y^{\text{clust}}_{i} \sim \mathrm{Gamma}\!\left(\mu^{\text{clust}}_{i}, \phi\right), \\
g\!\left(\mu^{m}_{i}\right)
&= \beta_0 + \beta_{1}I^{\text{napping}}_i + b_{\text{regime}(i)} + b_{\text{seed}(i)},
\end{align}
where $I^{\text{napping}}_i$ is the indicator of the napping type of regularization (with normalization being the reference), hence $\beta_0$ is the baseline corresponding to normalization and  $\beta_1$ is the fixed effect of napping regularization type; $b_{\text{regime}(i)} \sim \mathcal{N}(0, \sigma^2_{\text{regime}})$ is a per-regime random intercept; and $b_{\text{seed}(i)} \sim \mathcal{N}(0, \sigma^2_{\text{seed}})$ is a per-seed random intercept.


\section{Results}
Phase~1 isolates how napping duration and membrane noise shape performance within each regularization regime, and Phase~2 compares napping---fixed at its Phase~1 optimum---against weight normalization.
\subsection{Phase 1: Tuning napping}
\label{sec::res_phase1}

\subsubsection{Heatmaps}
\label{sec::res_heatmap}
\begin{figure}
    \centering
    \includegraphics[width=0.9\linewidth]{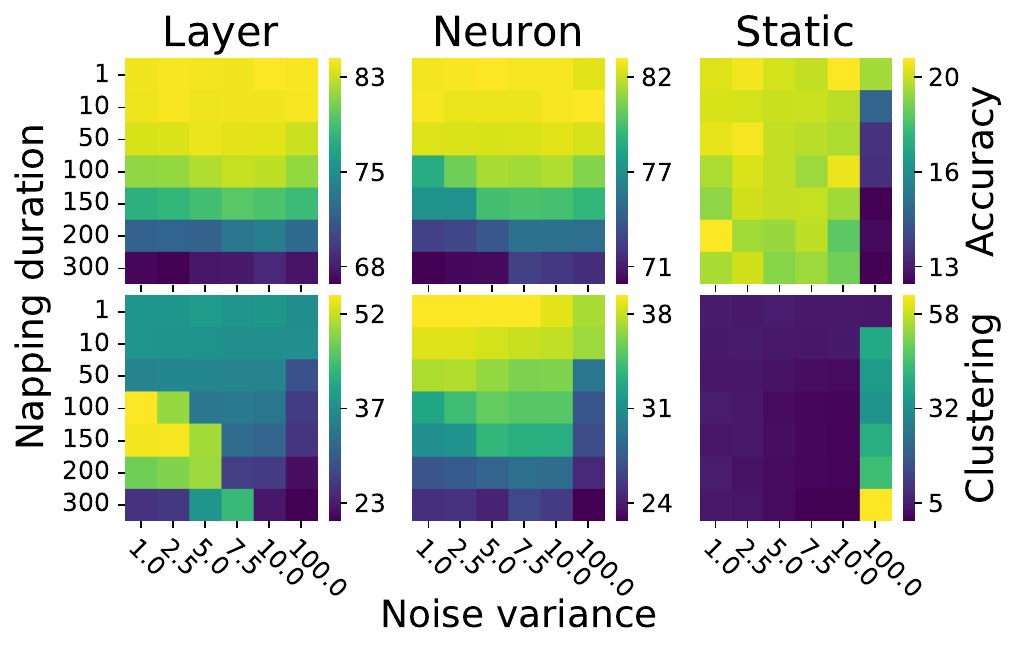}
    \caption{Predicted accuracy (top) and clustering (bottom) vs.\ duration and noise variance, across layer, neuron, and static regimes. Layer/neuron accuracy falls monotonically; layer clustering peaks mid-duration. Static stays flat except for anomalous high clustering at short duration, high noise ($\sigma^2 = 100$), with no accuracy gain.}
    \label{fig::res:heatmap}
\end{figure}

Accuracy and clustering peak in different regions of the duration--noise plane, and the two active regimes that accuracy treats as interchangeable separate clearly on clustering---see Figure~\ref{fig::res:heatmap}. The static regime stays near-flat on both metrics, consistent with its role as a weak reference, apart from an isolated high-clustering region at short duration and $\sigma^2 = 100$ that has no accuracy counterpart and suggests a qualitatively distinct representational regime under extreme noise.

\subsubsection{Mixed-effects models}
\label{sec::res_glmm}
The Beta GLMM (Table~\ref{tab:combined_glmm} and top panels of Figure~\ref{app::res_glmm_predicted}) quantifies the accuracy surface relative to the static regime reference. Both active regimes sit far above static on the logit scale (layer $\hat{\beta}_3 = +3.045$, neuron $\hat{\beta}_4 = +3.024$; both $p < .001$), and their estimates and confidence intervals are close enough that the two cannot be distinguished. Within the static reference the duration slope is weakly positive ($\hat{\beta}_1 = +0.024$, $p = .014$), but each active regime carries a strong negative napping duration interaction (layer $\hat{\beta}_6 =  -0.127$, neuron $\hat{\beta}_7 =  -0.117$; both $p < .001$) that reverses this trend, thus high-performing active regimes leverage shorter naps to sustain good accuracy. Noise does not have a significant main effect ($\hat{\beta}_2 = -0.020$, $p = .255$); its influence enters only through a small negative duration$\times$noise interaction ($\hat{\beta}_5 = -0.021$, $p < .001$) and modest positive three-way terms in the active regimes (layer $\hat{\beta}_{10} = +0.017$, $p = .007$; neuron $\hat{\beta}_{11}=+0.023$, $p < .001$), confirming that the high-noise column behaves slightly differently. Between-seed variance was negligible ($<.001$) implying that the results appear reproducible across seeds.

The Gamma GLMM for clustering (Table~\ref{tab:combined_glmm} and bottom panels of Figure~\ref{app::res_glmm_predicted}) separates the regimes somewhat more sharply than the Beta model, mirroring the clustering heatmap. Under the static reference, clustering falls steeply with duration ($\hat{\beta}_1 = -0.139$, $p < .001$), and both active regimes exceed the static method (layer $\hat{\beta}_3 =+1.515$, neuron $\hat{\beta}_4 =+1.594$; both $p < .001$). Their duration interactions diverge in a way that accuracy did not: the layer interaction ($\hat{\beta}_6 =+0.186$, $p < .001$) exceeds the static slope in magnitude, yielding a net positive duration effect for the layer regime---which corresponds to the ascending portion of the mid-duration peak in the heatmap---whereas the  neuron interaction was smaller ($\hat{\beta}_7 =+0.095$, $p < .001$) and leaves a net negative coefficient, consistent with the smooth neuron decline. Noise enters with a positive main-effect estimate ($\hat{\beta}_2 =+0.079$, $p = .006$), indicating higher clustering at greater noise, and a positive duration$\times$noise effect ($\hat{\beta}_5 = +0.072$, $p < .001$); the negative three-way estimates in both active regimes (layer $\hat{\beta}_{10} =-0.102$, neuron $\hat{\beta}_{11} =-0.077$; both $p < .001$) reduce the duration effect at high noise. Between-seed variance was $.003$.

Taken together, the two models tell a consistent story with one informative split: accuracy is governed almost entirely by regime and duration, with layer and neuron interchangeable and noise nearly inert, whereas clustering is also sensitive to noise---in addition to napping duration---and has layer and neuron as distinct regimes.

\footnotesize{
\begin{longtable}{lrrrl rrrl}

    \caption{Beta (accuracy) and Gamma (clustering) GLMM estimate effects.\label{tab:combined_glmm}}\\
    \toprule
    & \multicolumn{4}{c}{\textit{Beta (accuracy)}} & \multicolumn{4}{c}{\textit{Gamma (clustering)}}\\
    \cmidrule(lr){2-5}\cmidrule(lr){6-9}
    \textbf{Fixed effect} & Est. & \textit{z} & \textit{p} & & Est. & \textit{z} & \textit{p} & \\
    \endfirsthead
    \toprule
    & \multicolumn{4}{c}{\textit{Beta (accuracy)}} & \multicolumn{4}{c}{\textit{Gamma (clustering)}}\\
    \cmidrule(lr){2-5}\cmidrule(lr){6-9}
    Fixed effect & Est. & \textit{z} & \textit{p} & & Est. & \textit{z} & \textit{p} & \\
    \endhead
    $\hat{\beta}_0$ (intercept) & $-1.381$ & $-34.43$ & $<.001$ & *** & 2.114 & 30.24 & $<.001$ & ***\\[3pt]
    \multicolumn{9}{l}{\textit{Main effects}}\\
    $\hat{\beta}_1$ (dur) & .024 & 2.46 & $.014$ & * & $-.139$ & $-8.80$ & $<.001$ & ***\\
    $\hat{\beta}_2$ (noise) & $-.020$ & $-1.14$ & $.255$ & & .079 & 2.77 & $.006$ & **\\[3pt]
    $\hat{\beta}_3$ (layer) & 3.045 & 50.52 & $<.001$ & *** & 1.515 & 15.92 & $<.001$ & ***\\
    $\hat{\beta}_4$ (neuron) & 3.024 & 50.67 & $<.001$ & *** & 1.594 & 16.60 & $<.001$ & ***\\[3pt]
    \multicolumn{9}{l}{\textit{Two-way interactions}}\\
    $\hat{\beta}_5$ (dur$\times$noise) & $-.021$ & $-4.79$ & $<.001$ & *** & .072 & 10.69 & $<.001$ & ***\\
    $\hat{\beta}_6$ (layer$\times$dur) & $-.127$ & $-8.78$ & $<.001$ & *** & .186 & 8.06 & $<.001$ & ***\\
    $\hat{\beta}_7$ (neuron$\times$dur) & $-.117$ & $-8.11$ & $<.001$ & *** & .095 & 4.08 & $<.001$ & ***\\
    $\hat{\beta}_8$ (layer$\times$noise) & .041 & 1.57 & $.117$ & & $-.062$ & $-1.53$ & $.127$ &\\
    $\hat{\beta}_9$ (neuron$\times$noise) & .028 & 1.10 & $.273$ & & $-.085$ & $-2.06$ & $.039$ & *\\[3pt]
    \multicolumn{9}{l}{\textit{Three-way interactions}}\\
    $\hat{\beta}_{10}$ (layer$\times$dur$\times$noise) & .017 & 2.71 & $.007$ & ** & $-.102$ & $-10.41$ & $<.001$ & ***\\
    $\hat{\beta}_{11}$ (neuron$\times$dur$\times$noise) & .023 & 3.64 & $<.001$ & *** & $-.077$ & $-7.82$ & $<.001$ & ***\\
    \midrule
    \multicolumn{9}{l}{\textbf{Random effect variances}}\\
    Seed & \multicolumn{2}{r}{$<.001$} & & & \multicolumn{2}{r}{.003} & &\\
    \multicolumn{9}{l}{\textbf{Dispersion}}\\
    Model & \multicolumn{2}{r}{$\hat{\phi}=257$} & & & \multicolumn{2}{r}{$\hat{\sigma}^2=.070$} & &\\
    \bottomrule
    \caption*{Significance: * $p<.05$, ** $p<.01$, *** $p<.001$. $s$ = log napping duration (dur); $n$ = log noise variance. Reference regime = static.}
\end{longtable}
}

\subsection{Phase 2: Comparison with normalization}
\label{sec::res_phase2}
Phase~2 compares optimal napping against weight normalization directly, with regularization modes (static, neuron, layer) entering as random effects (Table~\ref{tab:phi_acc_glmm}). On accuracy, napping is statistically indistinguishable from normalization ($\hat{\beta}_{1} = 0.001$, $p = .936$). On clustering the two diverge: napping exceeds normalization by a small but significant margin ($\hat{\beta}_{1} = +0.156$, $p = .020$). The regime random intercept carries a large variance for accuracy ($1.930$), indicating that the regularization regime accounts for substantially more variation than the choice between napping and normalization. This is most likely due to the stark differences between the static regime compared to the two active regimes.

\footnotesize{
\begin{longtable}{@{}lrrrl@{\hspace{2.5em}}rrrl@{}}
    \caption{Beta (accuracy) and Gamma (clustering) GLMM results for the napping-vs-normalization comparison.\label{tab:phi_acc_glmm}}\\
    \toprule
    & \multicolumn{4}{c}{\textit{Beta (accuracy)}} & \multicolumn{4}{c}{\textit{Gamma (clustering)}}\\
    \cmidrule(lr){2-5}\cmidrule(lr){6-9}
    \textbf{Fixed effect} & Est. & \textit{z} & \textit{p} & & Est. & \textit{z} & \textit{p} & \\
    \endfirsthead
    \toprule
    & \multicolumn{4}{c}{\textit{Beta (accuracy)}} & \multicolumn{4}{c}{\textit{Gamma (clustering)}}\\
    \cmidrule(lr){2-5}\cmidrule(lr){6-9}
    \textbf{Fixed effect} & Est. & \textit{z} & \textit{p} & & Est. & \textit{z} & \textit{p} & \\
    \endhead
$\hat{\beta}_0$ (intercept) & \phantom{0}.566 & $\phantom{0}.71$ & $\phantom{<}.480$ &  & 3.188 & $8.37$ & $<.001$ & ***\\
$\hat{\beta}_1$ (napping) & .001 & $.08$ & $.936$ & & .156 & $2.33$ & $.020$ & *\\[3pt]
    \midrule
    \multicolumn{9}{l}{\textbf{Random effect variances}}\\
    Reg-mode & \multicolumn{2}{r}{1.930} & & & \multicolumn{2}{r}{.431} & &\\
    Seed & \multicolumn{2}{r}{$<.001$} & & & \multicolumn{2}{r}{$<.001$} & &\\[3pt]
    \multicolumn{9}{l}{\textbf{Dispersion and precision}}\\
    Model & \multicolumn{3}{r}{$\hat{\phi}=4.56\times10^{3}$} & & \multicolumn{3}{r}{$\hat{\sigma}^2=0.022$} & \\
    \bottomrule
    \caption*{Significance: * $p<.05$, ** $p<.01$, *** $p<.001$. Reference regularization: normalization, reg-mode (static/layer/neuron) as random intercept. $\hat{\phi}$ = Beta precision; $\hat{\sigma}^2$ = Gamma dispersion. Clustering model excludes the static 300 ms and $\sigma^2=100$ configuration as a pathological outlier (inflated clustering under collapsed accuracy).}
\end{longtable}
}

\section{Discussion}
\subsection{Main findings}
\label{sec::res_main_find}
Optimal napping is competitive with weight normalization: it matches normalization on accuracy and exceeds it on clustering. The two metrics, however, are optimized at different points of the duration--noise plane, and this divergence constrains how napping should be understood and deployed.

\subsubsection{Why accuracy and clustering diverge}
Accuracy peaks where proportional downscaling most closely tracks normalization---the shortest napping durations in the layer and neuron regimes---and degrades as duration grows, with membrane noise contributing only minor perturbations. Once napping is fixed at this optimum, it is statistically indistinguishable from normalization on accuracy.

Clustering follows a different pattern: for the layer and static regimes the strongest geometric separation arises at intermediate-to-long durations and moderate noise, precisely where accuracy is weakest. At its own optimum, napping exceeds normalization on clustering by a small but reliable margin, separating two regularizers that accuracy alone cannot distinguish. Gradual, noisy downscaling therefore confers a representational benefit that does not translate into linear decodability.

The two metrics constrain the representation differently---accuracy rewards \textit{linear decodability}, the Calinski--Harabasz index rewards label-free \textit{geometric separation}---and their optima need not coincide. In free-energy terms the split is the accuracy and complexity terms diverging: shorter, shorter, quieter naps track the data-fit that accuracy rewards, while longer, noisier ones favor the label-free geometry that clustering detects---consistent with, though not a direct measure of, lower model complexity. This is itself a reason to evaluate regularizers on the geometry of the learned representation rather than on a linear readout alone.

\subsection{Limitations}
Five limitations qualify the study. First, time constraints prevented full tuning of all network hyperparameters apart from the regularization sweep, so the results reflect napping in an unspecialized regime rather than its best case. We regard this as a deliberate trade-off: it offers an honest view of napping before the architecture is tailored to it, and since normalization operates under the same untuned conditions, the comparison remains on equal grounds. Second, we fixed two scheduling parameters that likely disadvantaged napping: the frequency of regularization, set to match our stable normalization schedule, and the weight-update interval, applied every timestep during napping but only every $100$~ms during ``wake,'' which likely injected excess noise into the napping phase. Both should themselves have been tuned. Third, the architecture may be too shallow to absorb noisy activity at scale during each napping episode, eroding the attractors that sustain class-specific features. Fourth, noisy relative downscaling may not be the most consequential aspect of sleep to implement; mechanisms such as memory replay, consolidation, slow-wave oscillations, or burst spiking patterns could matter more for improving stable representations. Finally, the method's runtime exceeds that of normalization and should demonstrate a clearer advantage before deployment at scale. Together these factors suggest our estimates are conservative and that work remains for the napping regularizer.

\subsection{Conclusion and Future Work}
Napping improves on prior periodic-sleep regularization in a comparable network---evidenced by the gap between our active regimes and the static reference---and is competitive with weight normalization, matching it on accuracy and surpassing it on clustering. The divergence between the two metrics motivates a potential future direction: fitting a dependent-mixture latent-state model over the population trajectories with a package such as \texttt{depmixS4} \cite{visser_depmixs4_2010}, and mapping inferred states onto digit labels---this would let napping and normalization be compared on the structure each induces, rather than on the linear separability a downstream classifier happens to exploit. The napping method is promising, particularly in clustering, but establishing a clear benefit across other networks and datasets remains open, and improving runtime performance presents an additional task to be completed.
\newpage

\bibliographystyle{splncs04}
\bibliography{references}
\newpage

\appendix
\section{Hyperparameters}
\label{sec::app:hyperparameters}
\begin{longtable}{m{3cm}m{2cm}m{4cm}m{2cm}}
\caption{Key tuning parameters for all experiments with \textit{biosnn} model}
\label{tab:hparams_snn_stdp}\\
\toprule
Component & Parameter & Descriptor & Value \\
\midrule
\endfirsthead

\toprule
Component & Parameter & Descriptor & Value \\
\midrule
\endhead

Network           &&&\\
                  & $N_{\text{in}}$ & Input neurons & 784 \\
                  & $N_{\text{exc}}$ & Excitatory neurons & 1024 \\
                  & $N_{\text{inh}}$ & Inhibitory neurons & 225 \\
                  & $P_{\text{in}\to\text{exc}}$ & Wiring prob. & 5\% \\
                  & $P_{\text{exc}\to\text{exc}}$ & Wiring prob. & 5\% \\
                  & $P_{\text{exc}\to\text{inh}}$ & Wiring prob. & 5\% \\
                  & $P_{\text{inh}\to\text{exc}}$ & Wiring prob. & 5\% \\
                  & $\mathcal{W}_{\text{in}\to\text{exc}}$  & Synaptic weight & 1.0 \\
                  & $\mathcal{W}_{\text{exc}\to\text{exc}}$ & Synaptic weight &  0.5\\
                  & $\mathcal{W}_{\text{inh}\to\text{exc}}$ & Synaptic weight & 1.0 \\
                  & $\mathcal{W}_{\text{exc}\to\text{inh}}$ & Synaptic weight & $-0.7$ \\
STDP              &&&\\
                  & $\eta$ & Learning rate & $4\times10^{-4}$ \\
                  & $\mu_{weight}$ & Weight dependence & $0.6$ \\
                  & $w_{max}$ & Max excitatory weight & $10$ \\
                  & $\tau_{trace}$ & Trace time constant & $20$ ms\\
                  & $\nu_{weight}$ & Weight update frequency & 100 ms\\
Neurodynamics   &&&\\
                  & $\Delta t$ & Time constant & 1 ms\\
                  & $\tau_m^{exc}$ & Membrane time constant & 20 ms \\
                  & $\tau_m^{inh}$ & Membrane time constant & 15 ms \\
                  & $\tau_{syn}^{exc}$ & Synapse current constant & 10 ms \\
                  & $\tau_{syn}^{inh}$ & Synapse current constant & 9 ms \\
                  & $U_{\text{rest}}$ & Resting potential & $-70$ mV \\
                  & $U_{\text{reset}}$& Reset potential  & $-80$ mV \\
                  & $U_{max}$ & Max potential & $40$ mV \\
                  & $U_{min}$ & Min potential & $-100$ mV \\
                  & $U_{th}$ & Spiking threshold & $-55$ mV \\
                  & $R_m$ & Membrane resistance & $15$ mV \\
                  & $\tau_{\text{th}}$ & Adaptive threshold constant & 200 ms \\
                  & $\delta$ & Threshold jump                & 0.5 mV\\
Regularization &&&\\
                  & $\nu_{reg}$ & Regularization frequency & 1050 ms\\
                  & $\omega$ & Napping duration & 100 ms\\ 
Data processing &&&\\
                  & $N_{\text{train}}$ & Training samples & 59000 \\
                  & $N_{\text{test}}$ & Test samples & 10000 \\
                  & $N_{\text{val}}$ & Validation samples & 1000 \\
                  & $S_{\text{image}}$ & Image dims & $(28, 28)$ \\
                  & $f_{\max}$ & Max spiking rate                 & 90 Hz \\
                  & $T_{\text{image}}$ & Stimulus duration         & 350 ms \\
                  & $B$ & Batch size & 1000 \\
                  & $N_{\text{batches}}$ & Number of batches & 60 \\
\bottomrule
\end{longtable}

\section{Neuron model}
\label{sec::app_neuron}
Each neuron integrates synaptic inputs via leaky integrate-and-fire 
(LIF) dynamics:
\begin{equation}
\tau_m \frac{dU_i(t)}{dt} = -(U_i(t) - U_{\text{rest}}) + 
R_m I_i^{\text{syn}}(t) + \xi_i(t),
\label{eq:lif}
\end{equation}
where $U_i(t)$ is the membrane potential, $\tau_m$ is the time 
constant, $R_m$ is the membrane resistance, $U_{\text{rest}}$ is the 
resting potential, $\xi_i(t) \sim \mathcal{N}(0,\sigma^2)$ is 
intrinsic noise, and $I^{\text{syn}}_i$ is the input current:
\begin{equation}
    I_i^{\text{syn}}(t) = \sum_j w_{ij} S_j(t),
\end{equation}
where $w_{ij}$ is the synaptic weight from neuron $j$ to neuron $i$, 
and $S_j(t) \in \{0,1\}$ indicates whether neuron $j$ fired at time 
$t$. When $U_i(t) \geq U_{\text{th}}^i(t)$, neuron $i$ emits a spike 
($S_i(t)=1$) and resets to $U_{\text{reset}}$. To prevent runaway 
activity, the threshold adapts as:
\begin{align}
    U_{\text{th}}^i(t) &= U_{\text{th}}(0)+\alpha_i, 
    \label{eq:threshold}\\
    \frac{d\alpha_i}{dt} &= -\frac{\alpha_i}{\tau_{\text{th}}}
    +S_i(t)\delta, \label{eq:alpha}
\end{align}
where $U_{\text{th}}(0)$ is the baseline threshold, $\alpha_i$ is a 
spike-driven adjustment that decays with time constant 
$\tau_{\text{th}}$ and increments by $\delta$ on each spike.

\section{Data}
\label{sec::app_data}
\subsection{Gabor filtering}
\label{sec::app:data:gabor}
We define the 2D Gabor filter as
\begin{equation}
    G(x,y;\theta) = 
    \exp\!\left(-\frac{x_\theta^2 + \gamma^2 y_\theta^2}
    {2\sigma^2}\right)
    \cos\!\left(\frac{2\pi x_\theta}{\Lambda}\right)
\end{equation}
where $x_\theta = x\cos\theta + y\sin\theta$, 
$y_\theta = -x\sin\theta + y\cos\theta$, $\sigma$ is the Gaussian 
envelope width, $\Lambda$ is the wavelength, and $\gamma=0.5$ is the 
spatial aspect ratio. Each filter response is rectified and 
normalized, then spatially rescaled into one quadrant of a composite 
image, preserving the original $28\times28$ resolution. This 
quadrant-packed image is passed directly to the spike encoder, 
yielding $N_{\text{in}} = 784$ input neurons.

\subsection{Data illustration}
\begin{figure}[H]
     \centering
     \begin{subfigure}[b]{0.3\textwidth}
         \centering
         \includegraphics[width=\textwidth]{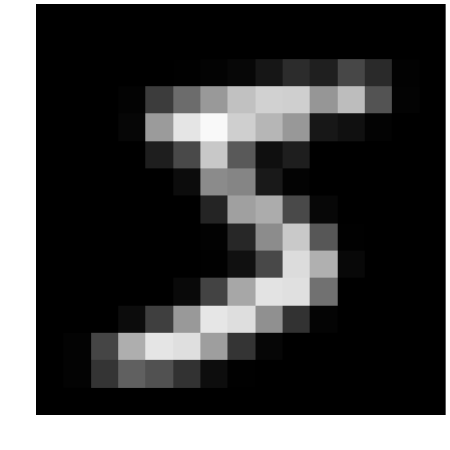}
         \caption{Raw digit}
         \label{fig:y equals x}
     \end{subfigure}
     \hfill
     \begin{subfigure}[b]{0.3\textwidth}
         \centering
         \includegraphics[width=\textwidth]{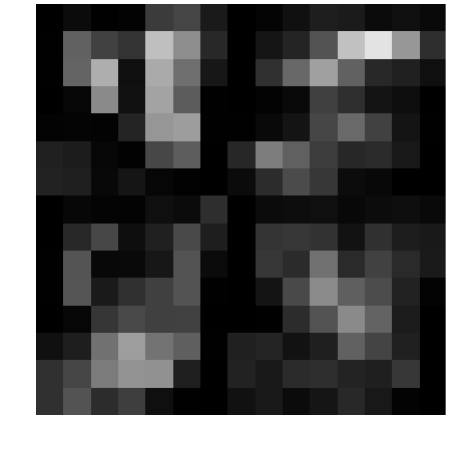}
         \caption{Gabor quadrants}
         \label{fig:three sin x}
     \end{subfigure}
     \hfill
     \begin{subfigure}[b]{0.30\textwidth}
         \centering
         \includegraphics[width=\textwidth]{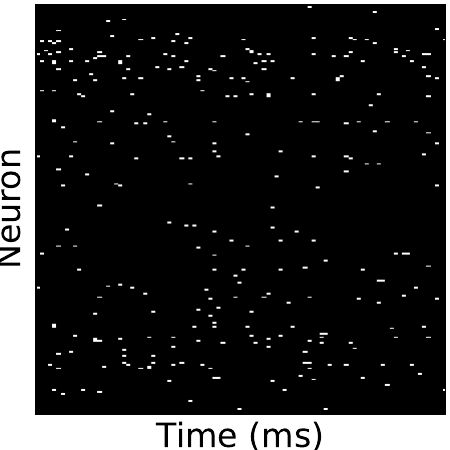}
         \caption{Spike raster}
         \label{fig:five over x}
     \end{subfigure}
        \caption{Illustration of data transformation pipeline. First, we extract MNIST digits, then we apply Gabor filters and obtain four quadrants reflecting each orientation, and finally, we convert the data to spike-encoded arrays.}
        \label{fig:three graphs}
\end{figure}

\section{GLMM}
\subsection{GLMM Phase~1}
Based on the obtained Phase~1 GLMM coefficients, we compute predictions across a grid of nap duration and noise levels to draw curves for each configuration. 
\begin{figure}
    \centering
    \includegraphics[width=0.8\linewidth]{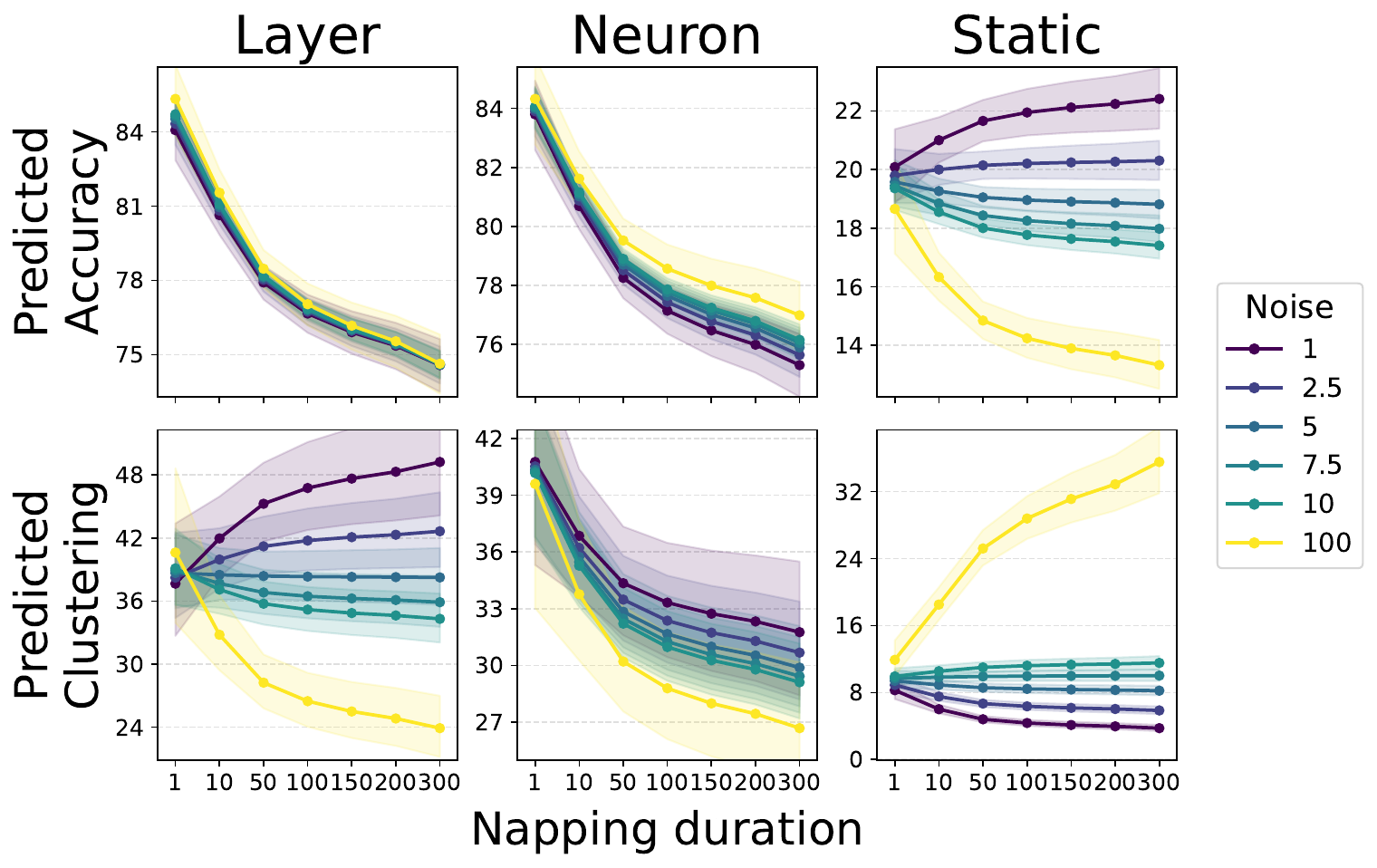}
    \caption{Beta GLMM predicted accuracy (top) and clustering (bottom) as a function of napping duration and noise level.}
    \label{app::res_glmm_predicted}
\end{figure}

We used the following nap duration and noise levels:
\begin{table}[H]
    \centering
    \begin{tabular}{lccccccc}
    \toprule
         Napping duration & 1 & 10 & 50 & 100 & 150 & 200 & 300 \\
         Noise level & 1.0 & 2.5 & 5.0 & 7.5 & 10 & 100 & \\
    \bottomrule
    \end{tabular}
    \caption{Due to the non-linear intervals between our selected values, we log-transformed the values in Phase~1 to ensure convergence of our GLMM.}
    \label{tab:app:glmm:ranges}
\end{table}

\subsection{GLMM Phase~2}
The optimal napping duration and noise level is estimated individually for each combination of regime and metric. All optimal setups per configurations are described in Table~\ref{tab::app:glmm:p2:optima}.

\begin{table}[H]
\centering
\caption{Accuracy- and clustering-optimal napping configurations (duration, noise variance) selected independently per regime in the Phase~2 sweep. Static remains near chance on accuracy, so its listed optima are included only for completeness. Its clustering optimum (300 ms, $\sigma^2 = 100$) was excluded from the Phase~2 comparison, as clustering inflates there while accuracy collapses to chance.}
\label{tab::app:glmm:p2:optima}
\begin{tabular}{l cc cc cc}
\toprule
& \multicolumn{2}{c}{\textit{Layer}} & \multicolumn{2}{c}{\textit{Neuron}} & \multicolumn{2}{c}{\textit{Static}}\\
\cmidrule(lr){2-3}\cmidrule(lr){4-5}\cmidrule(lr){6-7}
\textbf{Metric} & Dur. & Noise & Dur. & Noise & Dur. & Noise\\
\midrule
Accuracy   & 1 & 10 & 1 & 5 & 1 & 10\\
Clustering & 100 & 1 & 1 & 2.5 & 300 & 100\\
\bottomrule
\end{tabular}
\end{table}

\end{document}